\pdfoutput=1
\documentclass{article}
\usepackage{xcolor}
\usepackage{graphicx}
\usepackage{subcaption}
\usepackage{multirow}
\usepackage{amsthm}
\usepackage{amssymb}
\usepackage{enumitem}
\usepackage{microtype}
\usepackage{graphicx}
\usepackage{booktabs}
\usepackage{hyperref}

\usepackage[accepted]{mlsys2025}
\makeatletter
\renewcommand{\Notice@String}{Preprint.}
\makeatother
\hypersetup{pdfsubject={}} 
\usepackage{xspace}

\usepackage{amsmath,amsfonts,bm}

\def\eqref#1{equation~\ref{#1}}

\def\1{\bm{1}}

\DeclareMathAlphabet{\mathsfit}{\encodingdefault}{\sfdefault}{m}{sl}
\SetMathAlphabet{\mathsfit}{bold}{\encodingdefault}{\sfdefault}{bx}{n}

\usepackage{url}
\graphicspath{{figures/}}

\newtheorem{lemma}{Lemma}
\newtheorem{proposition}{Proposition}
\newtheorem{corollary}{Corollary}
\newtheorem{theorem}{Theorem}

\newcommand{\SystemName}{\textsc{FourSight}\xspace}
\newcommand{\splicegemm}{\textsc{SpliceGEMM}\xspace}

\newcommand{\PHB}[1]{\vspace{.2em}\noindent\textbf{#1}\hspace{.2em}}
\newcommand{\PHM}[1]{\vspace{.25em} \noindent\textbf{#1}\hspace{.2em}}

\mlsystitlerunning{\SystemName}

\begin{document}

\twocolumn[
\mlsystitle{\SystemName: Stable FP4 RL Post-Training via Sequence Filtering\\and Sparse Channel Compensation}

\mlsyssetsymbol{equal}{*}

\begin{mlsysauthorlist}
\mlsysauthor{Yining Wei}{equal,HKUST}
\mlsysauthor{Tianyuan Wu}{equal,HKUST}
\mlsysauthor{Gu Gong}{equal,HKUST-GZ}
\mlsysauthor{Yuechen Tao}{Alibaba}
\mlsysauthor{Yinghao Yu}{Alibaba}
\mlsysauthor{Xiaowen Chu}{HKUST-GZ}
\mlsysauthor{Wei Wang}{HKUST}
\end{mlsysauthorlist}
\mlsysaffiliation{HKUST}{HKUST}
\mlsysaffiliation{HKUST-GZ}{HKUST-GZ}
\mlsysaffiliation{Alibaba}{Alibaba Group}
\mlsyscorrespondingauthor{Tianyuan Wu}{twubt@connect.ust.hk}

\mlsyskeywords{Machine Learning, Reinforcement Learning, MLSys}
\vskip 0.3in

\begin{abstract}
Reinforcement learning (RL) post-training increasingly targets low-bit
deployment, with NVFP4 weight-and-activation quantization (W4A4) becoming the
state-of-the-practice format. However, integrating NVFP4 into RL degrades
rollout quality and amplifies cross-engine training--inference mismatch (TIM),
destabilizing training. We show that these failures are \emph{concentrated}
rather than uniform---severely mismatched tail tokens concentrate in a few outlier
sequences, while policy weights remain largely intact and quantization error
concentrates in a few high-variance activation channels. Guided by this, we
present \SystemName, an algorithm-system co-design with two targeted solutions:
TIM-aware sequence filtering that excludes drifted trajectories from policy
updates, and sparse precision compensation that keeps only top-variance channels
in BF16 while retaining native NVFP4 elsewhere. We further design a fused
\splicegemm kernel with specialized prefill and decode datapaths to minimize
overhead. \SystemName sustains near-BF16 accuracy and strong FP4 deployment
quality across models and engines, while improving step-to-accuracy by 1.76-2.60$\times$ and staying within 2.3\% of FP4 step
time up to 120B-scale models.

\end{abstract}
]

\printAffiliationsAndNotice{\mlsysEqualContribution}

\section{Introduction}
\label{sec:introduction}
Reinforcement learning (RL) post-training is now indispensable in the large
language model (LLM) lifecycle~\cite{deepseek2025r1}. To cut its steep compute
cost while optimizing models directly for low-precision serving, \emph{quantized
RL}~\cite{moonshot2025kimi,gu2026qarl,li2026qurl,zhuge2026quads} has gained
wide adoption, with flagship models increasingly post-trained~\cite{kimi2026k3}
or deployed~\cite{nvidia2026glm53flashnvfp4, nvidia2026qwen38nvfp4} in
aggressive low-bit formats such as \emph{NVFP4 weight-and-activation
quantization (W4A4)}. A typical NVFP4 RL pipeline pairs two decoupled engines: a
\emph{rollout engine}~\cite{kwon2023vllm,zheng2023sglang} running native NVFP4
GEMMs on accelerators such as NVIDIA Blackwell~\cite{nvidia2024blackwell} for
fast generation, and a \emph{training
engine}~\cite{shoeybi2019megatron,zhao2023fsdp,sheng2024hybridflow} performing
Quantization-Aware Training (QAT)~\cite{jacob2017quantization} in high precision
for gradient stability while using native or fake-quantized low-bit forwards to
match quantized rollout.

In practice, however, naive NVFP4 RL is unstable: even with truncated importance
sampling (TIS)~\cite{yao2025rollout}, training collapses midway and recovers
only partially (Figure~\ref{fig:obs_collapse_curves}). Two compounding failure
modes drive this breakdown. First, disparate numerical implementations across
rollout and training engines induce Training--Inference Mismatch
(TIM)~\cite{yao2025rollout}, where identical weights yield divergent
distributions. While manageable in high precision, this discrepancy is sharply
amplified by coarse low-bit representations, turning nominally on-policy
rollouts effectively off-policy and destabilizing
updates~\cite{yao2025offpolicy} (Figure~\ref{fig:obs_collapse_curves}). Second,
within the rollout stage, NVFP4 severely degrades sample quality. Across sequential
autoregressive decode passes, quantization error at each step feeds into the
next and accumulates, yielding drifted rollouts that starve policy updates of
effective learning signals~\cite{mak2026hifloat4,zhuge2026quads}.

Existing remedies for these failure modes force a costly trade-off between
training stability and system efficiency. Algorithmically, TIM corrections clip
or mask individual tokens~\cite{yao2025rollout,li2026qurl,lingteam2025stepevolves}
or mask entire sequences~\cite{gu2026qarl,li2026trm}, yet both fall short in
NVFP4 RL. Token-level schemes still leak distorted sequence-level advantages via
unclipped tokens, whereas sequence-level filters like QaRL~\cite{gu2026qarl}
rely on coarse aggregated metrics that miss intra-sequence breakdowns, hurting
learning efficiency and leaving degraded rollouts uncompensated
(Figure~\ref{fig:main_new_stack}).
At the operator level, precision compensations such as
ARCQuant~\cite{meng2026arcquant}, HiFloat4~\cite{mak2026hifloat4}, and
QUADS~\cite{zhuge2026quads} restore rollout quality via re-quantization or
mixed-precision compute, but add overhead that largely offsets low-bit
throughput gains. At the system level, runtime unification like
veXact~\cite{zhong2026vexact} eliminates TIM by forcing both stages onto a
shared engine, but sacrifices stage-specific optimizations at a $\sim 25.3\%$ RL
step slowdown~\cite{jiang2026isoexec}.

We show that this trade-off is avoidable because NVFP4 RL failures are
\emph{concentrated} rather than uniform, a skewness obscured by aggregate
metrics like entropy and cross-engine KL that spike only after
collapse (Figure~\ref{fig:obs_collapse_curves}). At the token level, severe
mismatch is driven by nonsensical \emph{tail tokens} sampled with high
probability during rollout but assigned near-zero training probability
(Figure~\ref{fig:obs_offpolicy_scatter}). In logit space, they form a distinct
off-diagonal cluster while other tokens stay aligned
(Figure~\ref{fig:obs_logit_scatter}). Instead of spreading evenly across a
batch, tail tokens are rare before collapse (0.2\% in a batch), concentrate in a
subset of trajectories at the collapse trough (361/1024 in our experiments), and
polarize further after the partial rebound, where nearly 89\% of trajectories
recover while the rest consist almost entirely of tail tokens
(Figure~\ref{fig:obs_tail_trajectories}). Because GRPO-style algorithms give
every token of a trajectory the same advantage, clipping the tail tokens still
lets the rest of derailed trajectories update the policy, so an effective defense
must discard whole trajectories.

Rollout degradation is likewise concentrated: it stems from quantization
errors in a small fraction of activation channels rather than corrupted
weights. In our experiment, switching only the rollout precision of a collapsed Qwen3-8B FP4
checkpoint to BF16 recovers 20~pp of GSM8K test accuracy
(Figure~\ref{fig:obs_precision_rescue}a). Across training, weights drift by at
most 0.11\% while relative policy quantization error grows by nearly 8$\times$
(Figure~\ref{fig:obs_precision_rescue}c), showing that RL leaves weights largely
intact but amplifies quantization sensitivity. Consistent with prior reports of
activation outliers in
LLMs~\cite{dettmers2022gpt3,xiao2023smoothquant,meng2026arcquant}, this error
concentrates in a few high-variance input channels
(Figure~\ref{fig:obs_precision_rescue}b). Under NVFP4, each such channel also
inflates the scale shared by its 16-element block and coarsens the other 15
values~\cite{mak2026hifloat4,zhuge2026quads}, so compensating only these
channels recovers most of the error.

Guided by these observations, we develop \SystemName, an algorithm--system
co-design with two complementary components. First, \emph{TIM-aware sequence
filtering} zeroes out the advantage of drifted trajectories to exclude them from
policy updates. Since the rollout engine sampled every trajectory token, a
training probability below a small threshold directly signals a severe mismatch;
we mark such tokens as \emph{tail tokens}. \SystemName discards a trajectory if
it contains many tail tokens in total (widespread numerical drift) or a long run
of consecutive tail tokens (autoregressive generation collapse), while
tolerating rare, isolated tail tokens in healthy rollouts. Second, to restore
rollout quality, \SystemName applies \emph{sparse precision compensation} before
each rollout stage, keeping the highest-variance input channels of each
projection (top-64 in our experiments) in BF16 and the rest in native NVFP4. To
avoid separate residual paths~\cite{mak2026hifloat4,zhuge2026quads}, we design
\splicegemm, a fused kernel that folds this BF16 correction into the NVFP4 GEMM
with specialized datapaths for large-batch prefill and training and small-batch
decoding (Section~\ref{sec:impl}). Together, compensation supplies
higher-quality rollouts to learn from, while filtering removes the residual
drift it leaves behind.

We have implemented \SystemName atop veRL~\cite{sheng2024hybridflow}, using vLLM
for rollout and Megatron-LM or FSDP for training. \SystemName closely tracks
BF16 validation accuracy on both Qwen3-8B with DAPO and Qwen3-1.7B with GRPO,
and on Qwen3-8B reaches 20\% AIME24 accuracy in 1.76--2.60$\times$ fewer steps
than QaRL, the strongest quantized baseline (Figure~\ref{fig:main_new_stack}).
When served in FP4 with sparse compensation, \SystemName outperforms
BF16-trained models by 12.7--16.4~pp on Qwen3-1.7B and matches or exceeds all
quantized baselines on nine of ten benchmark--engine pairs on Qwen3-8B. System
evaluations from Qwen3-8B to Nemotron3-Super-120B-A12B on 32~GPUs show that
\SystemName stays within 2.3\% of uncompensated FP4 step time, running up to
1.40$\times$ faster than BF16 and 1.18--1.29$\times$ faster than QUADS. Finally,
ablations confirm that both components are complementary and that protecting 64
out of 4096 variance-selected channels per projection recovers most of the FP4
accuracy loss on a degraded checkpoint.

\section{Background}
\label{sec:background}

\subsection{Quantized RL Post-training}
\label{subsec:bg_fp4}

\PHB{Low-precision RL with NVFP4.} RL post-training iterates over two decoupled
stages~\cite{sheng2024hybridflow}: a \emph{rollout stage} that autoregressively
generates response trajectories using an inference
engine~\cite{kwon2023vllm,zheng2023sglang}, and a \emph{training stage} that
computes policy gradients and updates model parameters using a distributed
training engine~\cite{shoeybi2019megatron,zhao2023fsdp}. Applying low-bit
quantization to this RL pipeline both accelerates post-training and directly
adapts models for low-precision deployment~\cite{moonshot2025kimi, li2026qurl},
where the rollout stage executes in low precision for efficiency while the
training stage adopts Quantization-Aware Training
(QAT)~\cite{jacob2017quantization} for gradient stability. Aggressive
quantization with NVFP4~\cite{nvidia2024blackwell} has emerged as the \emph{de facto}
standard of RL. NVFP4 overcomes the narrow dynamic range of raw FP4 (E2M1) by
combining a tensor-wise FP32 global scale with hierarchical block scaling, and
partitioning tensors along the reduction dimension into 16-element blocks that
each shares a local FP8 (E4M3) scale. However, this block-wise sharing is
vulnerable to outliers: if a block contains even a single large outlier, its
shared local scale is inflated, severely coarsening the quantization resolution
for the remaining 15 elements in that block~\cite{mak2026hifloat4,
zhuge2026quads}.

\PHM{Group-relative policy optimization (GRPO).}
Modern RL post-training widely adopts group-relative algorithms such as GRPO~\cite{shao2024deepseekmath} and DAPO~\cite{yu2025neurips-dapo}.
For each prompt $x$, the rollout policy $\theta_{\text{old}}$ samples a group of $G$ trajectories $\{y_i\}_{i=1}^G$, each associates with a reward $r_i$. The advantage $\hat{A}_i$ is calculated by $\hat{A}_i = (r_i - \mu)/\sigma$, where $\mu,\sigma$ is the mean and std of the group's rewards. The objective of GRPO is
\begin{equation}
\label{eq:grpo}
\mathcal{L}_{\mathrm{GRPO}}(\theta)
=
\mathbb{E}\!\left[
\frac{1}{G}\sum_{i=1}^{G}\frac{1}{|y_i|}\sum_{t=1}^{|y_i|}
\ell_{i,t}^{\mathrm{clip}}(\theta)
\right],
\end{equation}
where $\ell_{i,t}^{\mathrm{clip}}(\theta) = \min\bigl(\rho_{i,t}\hat{A}_i, \, \mathrm{clip}(\rho_{i,t}, 1-\epsilon, 1+\epsilon)\hat{A}_i\bigr)$ denotes the per-token PPO-clipped surrogate objective~\cite{schulman2017proximal} with clipping threshold $\epsilon$, and $\rho_{i,t} = \pi(y_{i,t} \mid y_{i,<t}; \theta) /\pi(y_{i,t} \mid y_{i,<t}; \theta_{\mathrm{old}})$ is the token-level importance sampling ratio evaluated under the policy $\pi$.

\subsection{TIM and Existing Remedies}
\label{subsec:bg_rl_tim}
\PHB{Training--inference mismatch (TIM).}
In principle, the rollout policy $\hat{\pi}_{\mathrm{roll}}(\cdot \mid \theta_{\mathrm{old}})$ generating trajectories and the training policy $\pi_{\mathrm{train}}(\cdot \mid \theta_{\mathrm{old}})$ evaluating the importance ratio $\rho_{i,t}$ represent the identical model $\theta_{\mathrm{old}}$.
In practice, however, the decoupled engines employ disparate kernel implementations, parallel reduction orders, and quantization execution modes, causing their evaluated token probabilities to diverge---a phenomenon known as \emph{Training--Inference Mismatch (TIM)}~\cite{yao2025rollout, gu2026qarl}:
\begin{equation}
    \hat{\pi}_{\mathrm{roll}}(y_{i,t} \mid y_{i,<t}; \theta_{\mathrm{old}}) \neq \pi_{\mathrm{train}}(y_{i,t} \mid y_{i,<t}; \theta_{\mathrm{old}}).
    \label{eq:tim}
\end{equation}
Under aggressive NVFP4 quantization, coarse 4-bit rounding and outlier-induced block-scale inflation severely amplify this cross-engine numerical divergence over long reasoning sequences.
Because trajectories are sampled from $\hat{\pi}_{\mathrm{roll}}$ while policy gradients are evaluated solely under $\pi_{\mathrm{train}}$, severe TIM turns nominally on-policy updates into uncalibrated off-policy updates, injecting biased gradients that destabilize RL training.

\PHB{Existing remedies and limitations.}
To mitigate this mismatch, prior works widely adopt algorithmic corrections, most notably Truncated Importance Sampling (TIS)~\cite{yao2025rollout, li2026qurl}, which reweights each token's surrogate objective $\ell_{i,t}^{\mathrm{clip}}(\theta)$ by a clipped importance ratio $\min\bigl(c, \frac{\pi_{\mathrm{train}}(y_{i,t} \mid y_{i,<t}; \theta_{\mathrm{old}})}{\hat{\pi}_{\mathrm{roll}}(y_{i,t} \mid y_{i,<t}; \theta_{\mathrm{old}})}\bigr)$ with threshold $c > 1$.
However, under aggressive NVFP4 quantization, such token-level ratio clipping---as well as token-level or heuristic masking~\cite{lingteam2025stepevolves, gu2026qarl, li2026trm}---merely suppresses individual mismatched tokens without isolating corrupted trajectories, allowing distorted gradients to leak through the shared sequence-level advantage $\hat{A}_i$ and eventually trigger policy collapse.
Alternatively, system- and operator-level remedies seek to reduce numerical mismatch at the source, either by compensating quantization errors via partial high-precision recomputation~\cite{meng2026arcquant, mak2026hifloat4, zhuge2026quads} or by forcing both stages onto a unified runtime engine~\cite{zhong2026vexact}, yet both sacrifice low-bit or stage-specific optimizations and incur severe slowdowns.

\section{Observations and Motivation}
\label{sec:obs}

\begin{figure}[!tb]
    \centering
    \includegraphics[width=\linewidth]{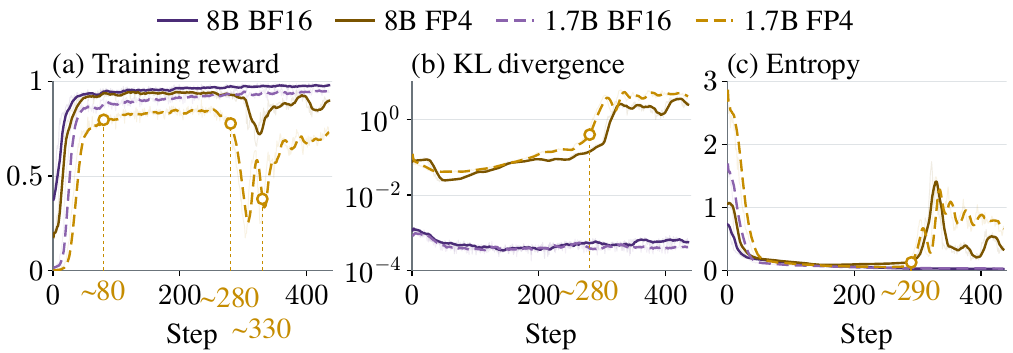}
    \caption{Macroscopic training dynamics of Qwen3-8B and Qwen3-1.7B under BF16 and native FP4 (with default TIS) on GSM8K~\cite{cobbe2021gsm8k} dataset: (a) mean training reward, illustrating rapid initial learning followed by mid-training collapse ($\sim$Step 280) and an incomplete, fluctuating rebound ($\sim$Step 330); (b) sampled train--inference KL divergence; and (c) policy entropy.}
    \label{fig:obs_collapse_curves}
\end{figure}

We argue that enabling stable and efficient quantized RL must overcome two
challenges: (i) \emph{cross-stage TIM}, where coarse NVFP4 representations
sharply exacerbate TIM and cause severe policy drift. The drifted gradients then
accumulate through the RL loop and eventually manifest as catastrophic
degradation in later training steps~\cite{mak2026hifloat4, zhuge2026quads}; (ii)
\emph{within-stage rollout degradation}, where the quantization error is
amplified through thousands of autoregressive decoding passes, leading to
low-quality rollouts that starve policy updates of effective learning
signals~\cite{yao2025rollout}. 

Figure~\ref{fig:obs_collapse_curves} shows their impact on Qwen3-1.7B and 8B
training with TIS correction already enabled. First, although both precisions
plateau around Step 80, FP4 consistently lags behind BF16 due to degraded
rollout quality (Figure~\ref{fig:obs_collapse_curves}a). Second, while BF16
remains stable, FP4 abruptly collapses around Steps 280--330 as TIM accumulates
over training, followed by only a partial rebound at Step 330. Third, aggregate
policy KL divergence and entropy (Figures~\ref{fig:obs_collapse_curves}b, c) are
poor predictors: their spikes lag behind the collapse and fail to track the
subsequent reward rebound. We therefore look beyond aggregate metrics to analyze
finer-grained TIM (Section~\ref{sec:obs_tim}) and channel-level rollout
sensitivity (Section~\ref{sec:obs_rollout}).

\subsection{Catastrophic TIM Concentrates in Outlier Trajectories}
\label{sec:obs_tim}

In this section, we consider inspecting into the token-level TIM of the Qwen3-1.7B training on GSM8K dataset, because the batch- or sequence-aggregated metrics mask potential localized failure modes within a trajectory.

\begin{figure}[!tb]
    \centering
    \includegraphics[width=\linewidth]{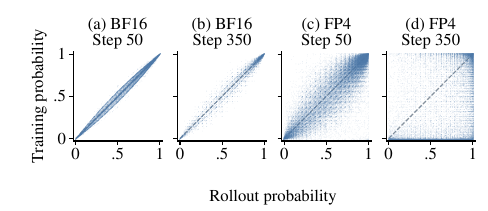}
    \vspace{-0.5cm}
    \caption{Token-level rollout probability ($\hat{\pi}_{\mathrm{roll}}$) versus training probability ($\pi_{\mathrm{train}}$) of Qwen3-1.7B at Step 50 and Step 350. While BF16 remains tightly aligned along $y=x$ (a, b), FP4 evolves from a wider fusiform envelope (c) to severe dispersion with dense accumulation along the $\pi_{\mathrm{train}} \approx 0$ boundary (d).}%
    \label{fig:obs_offpolicy_scatter}
    \vspace{-0.5cm}
\end{figure}

\PHB{Token-level probability divergence and boundary accumulation.}
Figure~\ref{fig:obs_offpolicy_scatter} illustrates the scatter diagrams $(\hat{\pi}_{\mathrm{roll}}, \pi_{\mathrm{train}})$ of each token in a batch. 
The scatter plots of the BF16 experiment (Figure~\ref{fig:obs_offpolicy_scatter}a,b) manifest a relative ideal on-policy regime, where points concentrate tightly to a narrow fusiform envelope along the identity diagonal $y=x$. In native FP4 setting, however, quantization noise noticeably widens the envelope even during the healthy early stage (Step 50, Figure~\ref{fig:obs_offpolicy_scatter}c). By Step 350 (Figure~\ref{fig:obs_offpolicy_scatter}d), the joint distribution deteriorates dramatically: points appear widely scattered across the $[0, 1]^2$ plane, and a dense horizontal band accumulates along the bottom boundary $\pi_{\mathrm{train}} \approx 0$. In this region, tokens sampled with high confidence by the rollout engine are assigned near-zero probability by the training engine, representing extreme off-policy divergence.

\begin{figure}[!tb]
    \centering
    \includegraphics[width=\linewidth]{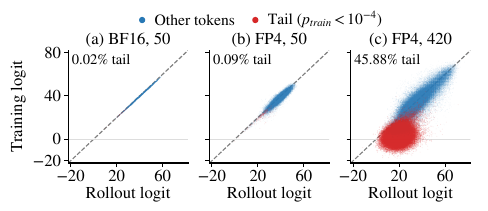}
    \caption{Pre-softmax rollout logit ($z_{\mathrm{roll}}$) versus training logit ($z_{\mathrm{train}}$) of Qwen3-1.7B. Unwrapping Softmax reveals that at Step 420, these tail tokens form a distinct off-diagonal cluster, with moderate rollout logits but training logits $z_{\mathrm{train}}\approx 0$.}%
    \label{fig:obs_logit_scatter}
    \vspace{-0.4cm}
\end{figure}

\PHM{The separable off-diagonal tail cluster in logit space.} While Figure~\ref{fig:obs_offpolicy_scatter}d suggests widespread scatter across $[0, 1]^2$, the exponential Softmax mapping both exaggerates minor perturbations among top-ranked logits in the interior and compresses all small probabilities ($\pi_{\mathrm{train}} \in [10^{-10}, 10^{-2}]$) onto the same $\pi_{\mathrm{train}} \approx 0$ boundary. To isolate true divergence from Softmax distortion, we inspect the pre-softmax logits $(z_{\mathrm{roll}}, z_{\mathrm{train}})$ in Figure~\ref{fig:obs_logit_scatter}.
This view resolves the apparent global chaos of Figure~\ref{fig:obs_offpolicy_scatter}d: even long after collapse (e.g., step 420 in Figure~\ref{fig:obs_logit_scatter}c), over a half of FP4 tokens (blue) remain well-aligned within a fusiform envelope. The severe mismatch is instead driven by a localized \emph{off-diagonal anomaly cluster} breaking away below the diagonal, where tokens maintain moderate-to-high rollout logits yet collapse toward zero or negative training logits ($z_{\mathrm{train}} \approx 0$). Crucially, highlighting tokens that fall into the extreme tail of the training distribution ($\pi_{\mathrm{train}}(y_{i,t} \mid y_{i,<t}) < \tau = 10^{-4}$, marked as \emph{Tail token} in red) cleanly isolates this off-diagonal logit cluster. Decoding these tail tokens reveals that they correspond to semantically degenerate outputs---such as garbled subwords, repetitive symbol loops, and abrupt cross-lingual drift---hallucinated by the low-bit rollout engine but rejected by the training engine.

\begin{figure}[!tb]
    \centering
    \includegraphics[width=\linewidth]{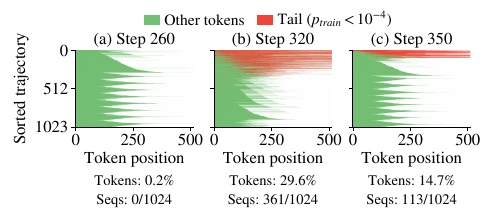}
    \caption{Spatial distribution of tail tokens ($\pi_{\mathrm{train}} < 10^{-4}$, red) versus normal tokens (green) of Qwen3-1.7B across 1,024 batch trajectories sorted by tail-token fraction at Step 260, 320, and 350.}
    \label{fig:obs_tail_trajectories}
    \vspace{-0.4cm}
\end{figure}

\PHM{Trajectory-level concentration and why token-level TIS fails.}
To understand how this off-diagonal cluster destabilizes optimization, Figure~\ref{fig:obs_tail_trajectories} visualizes the positions of tail tokens (red) in a training batch at three representative steps: (1) \textit{Pre-collapse (Step 260, Figure~\ref{fig:obs_tail_trajectories}a):} Tail tokens constitute only $0.2\%$ of generated tokens and randomly occur as isolated, sporadic points
(2) \textit{Collapse trough (Step 320, Figure~\ref{fig:obs_tail_trajectories}b):} The tail-token fraction surges to $29.6\%$. Rather than distributing uniformly across the batch, tail tokens cluster heavily within $361/1024$ ($35.0\%$) trajectories. 
(3) \textit{Partial rebound (Step 350, Figure~\ref{fig:obs_tail_trajectories}c):} The concentration becomes even more polarized. Nearly $89.0\%$ of trajectories return to healthy generation (green) while the remaining outlier trajectories are almost completely saturated with tail tokens, which explains why the batch-mean training reward partially rebounds in Figure~\ref{fig:obs_collapse_curves}a. This skewed distribution explains why standard TIS cannot prevent policy collapse. In GRPO or DAPO, the sequence level advantage 
still contain the information from the clipped corrupted token. 
The concentration of these anomalous tokens inspires that suppressing their impact requires gating credit assignment at the \emph{sequence level} rather than clipping tokens in isolation.

\subsection{Rollout Degradation Stems from Quantization Sensitivity rather than Corrupted Weights}
\label{sec:obs_rollout}

In this section, we focus on the numerical behaviors behind rollout degradation and explore the sparsity of quantization errors based on the Qwen3-8B model trained on GSM8K.

\begin{figure}[!tb]
    \centering
    \includegraphics[width=\linewidth]{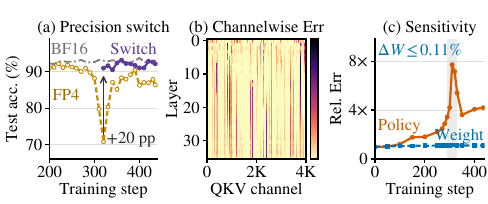}
    \caption{Diagnosis of Qwen3-8B rollout degradation: (a) switching rollout from FP4 to BF16 immediately recovers $+20$\,pp GSM8K test accuracy; (b) layer-wise and channel-wise quantization error across QKV projections, showing concentration in a few outlier channels; and (c) evolution of relative policy output error versus relative weight change across training steps.}
    \label{fig:obs_precision_rescue}
\end{figure}

\PHB{Intact weights versus amplified quantization error.}
To diagnose whether the mid-training collapse reflects irreversible corruption of the learned model weights or a failure of low-precision generation, we take the degraded FP4 checkpoint and switch \emph{only} the rollout precision from W4A4 to BF16 (Figure~\ref{fig:obs_precision_rescue}a). Remarkably, this switch yields an immediate $+20$\,pp surge in test accuracy, closely approaching the accuracy trajectory of a model trained in BF16 from scratch. Furthermore, Figure~\ref{fig:obs_precision_rescue}c shows the relative parameter drift alongside the model's output policy error compared to BF16 across training steps: while the policy weights change minimally throughout training ($\|\Delta W\|_2 \le 0.11\%$), the relative policy error induced by W4A4 quantization escalates by nearly $8\times$. These results confirm that \emph{the weights remain largely intact even at the lowest point of collapse, while RL updates progressively amplify the rollout policy's sensitivity to quantization noise}.

\PHM{The sparsity of quantization error.}
We further investigate the source of quantization error at a per-layer and per-channel granularity.
As shown in Figure~\ref{fig:obs_precision_rescue}b, severe quantization errors consistently concentrate within a tiny fraction of outlier channels with disproportionately large magnitudes or variance across layers, aligning with the findings of prior studies about massive activations and quantization outliers~\cite{dettmers2022gpt3, xiao2023smoothquant, meng2026arcquant}. NVFP4's block scaling mechanism amplifies this error, because the outlier channel inflates the shared local scale factor, coarsening the quantization grid for the remaining 15 elements in the same block~\cite{mak2026hifloat4, zhuge2026quads}. This sparsity inspires us that protecting only the outlier channels can compensate most of error. 

\section{\SystemName Algorithm Design}
\label{sec:design}

\SystemName enables stable and efficient FP4 RL post-training through algorithm-system co-design. It comprises two key components: (1) a TIM-aware sequence filtering algorithm that discards trajectories containing concentrated large-TIM tail tokens to prevent corrupted gradients from destabilizing policy updates, and (2) a sparse precision compensation mechanism that restores rollout quality by selectively executing high-variance activation channels in BF16 while retaining native NVFP4 for the vast majority of computations. We detail the algorithmic design of \SystemName below and present its system implementations in Section~\ref{sec:impl}.

\subsection{TIM-aware Sequence Filtering}

\PHB{Design principles.} Since severe TIM is concentrated within outlier trajectories and contaminates their shared advantage (Section~\ref{sec:obs_tim}), adopting token-level regularization (e.g., TIS) alone has proven ineffective. Therefore, \SystemName's filtering algorithm operates at the sequence level to isolate drifted trajectories from the policy update; yet, since normal trajectories occasionally exhibit isolated large-TIM tokens, the filtering is designed to be robust to sporadic token jitter to avoid discarding valid sequences.

\PHM{The dual-gate mechanism.}
We mark a token as a \emph{tail token} if its training probability $\pi_{\text{train}}(y_{i,t}) \le \tau$, where $\tau$ is a small constant defining the threshold. These tail tokens correspond to the boundary-accumulated tokens in Figure~\ref{fig:obs_offpolicy_scatter} and the off-diagonal cluster in Figure~\ref{fig:obs_logit_scatter}. To prevent large-TIM sequences from entering the policy update, \SystemName uses a dual-gate mechanism to filter sequences according to this definition. (1) \texttt{G1} (Cumulative count): A trajectory is filtered out if it contains at least $K_1$ tail tokens in total. A high volume of tail tokens indicates widespread numerical drift across the trajectory; discarding such trajectories preserves overall gradient hygiene. (2) \texttt{G2} (Consecutive count): A trajectory is also filtered out if it contains at least $K_2$ consecutive tail tokens ($K_2 \le K_1$). Because autoregressive decoding conditions on prior tokens, a consecutive run of tail tokens signals that the rollout has entered a degenerate prefix and suffered localized mode collapse.

\PHM{GRPO objective with filtering.} Define the filter $f_i$ as
\begin{equation}
    f_i = \mathbb{I}\{\mathrm{G1}(y_i, \tau) < K_1 \land \mathrm{G2}(y_i, \tau) < K_2\}.
\end{equation} 
A sequence $y_i$ participates in gradient calculation only when $f_i\ne 0$. We apply filtering post hoc by directly masking the sequence advantage:
$\hat{A}'_i = f_i \hat{A}_i = f_i \frac{r_i - \mu}{\sigma}$, yielding the corrected GRPO objective:
\begin{equation}
\mathcal{L}_{\mathrm{GRPO}}(\theta) = \mathbb{E}\left[ \frac{1}{G} \sum_{i=1}^{G} \frac{1}{|y_i|} \sum_{t=1}^{|y_i|} \ell_{i,t}^{\mathrm{clip}}(\theta; \hat{A}'_i) \right].
\label{eq:sf-grpo}
\end{equation}
Crucially, we preserve the original denominator $G$ 
rather than renormalizing by the surviving sequence count $\sum_{i=1}^G f_i$, because reducing the denominator amplifies the gradient signals contributed by unfiltered sequences and implicitly increases the learning rate.

\subsection{Sparse Precision Compensation}
Besides sequence filtering, \SystemName uses sparse precision compensation to improve rollout quality. The key insight is that model weights are insensitive to quantization throughout training, whereas the activation outliers that degrade rollouts are concentrated in only a few channels (Figure~\ref{fig:obs_precision_rescue}). In these channels, activations exhibit large variance: a single outlier amplifies its block's scaling factor and thus affects its neighboring values. Since the vast majority of channels are benign to quantization, compensation only on these sparse outlier channels can preserve most of numerical precision.

\PHM{Mix-precision GEMM decomposition.} \SystemName implements this sparse compensation by the \splicegemm operator. For a linear layer \(Y=XW^\top\) with \(X\in\mathbb{R}^{M\times K}\) and \(W\in\mathbb{R}^{N\times K}\), \splicegemm partitions the \(K\) input channels into a sparse set \(S\) and its complement \(\bar S\). The main product uses quantized NVFP4 operands on \(\bar S\), while the channels in \(S\) are kept unquantized and applied as a BF16 correction:
\begin{equation}
Y \approx \alpha\left(\widehat X_{\bar S}\widehat W_{\bar S}^{\top}
                      + X_S {W'_S}^{\top}\right),
\label{eq:splicegemm}
\end{equation}
where $W'_S=\operatorname{BF16}(W_S/\alpha)$ is the dequantized BF16 weight, $\widehat X_{\bar S}$ and $\widehat W_{\bar S}$ denote the NVFP4-quantized weights and activations, $\alpha$ is the global NVFP4 quantization scale, and $X_S$ retains its original BF16 values.

\PHM{Variance-based channel selection.} \SystemName protects the top-$s$ channels with the largest activation variance, where $s$ is a predefined channel budget. We prove that this selection policy effectively bounds the quantization error under a fixed budget $s$: suppose we protect an $s$-element channel subset $S$, and let $\widehat y_S$ and $y = Wx$ denote the compensated and original BF16 outputs, respectively. The expected reconstruction error is upper-bounded by:
\begin{equation}
    \mathbb{E}\lVert\widehat y_S-y\rVert_2^2
  \le C\lVert W\rVert_F^2\sum_{j\notin S}m_j\approx C\lVert W\rVert_F^2\sum_{j\notin S}\sigma_j^2,
\end{equation}
where $m_j = \mathbb{E}[x_j^2]$ is the second moment of channel $j$ and $C = \left[\sqrt{B}\,c_W + \sqrt{B}\,c_A\left(1+\sqrt{B}\,c_W\right)\right]^2$ . Under NVFP4 quantization with block size $B=16$ and scale-rounding constants $c_A=c_W=17/96$, we have $C\approx 3.68$.

The full proof is in Appendix~\ref{append: theory_proof}. This bound indicates that the output quantization error is governed by the second moments of the unprotected channels ($j \notin S$). Crucially, when channel means are small, second moments are well approximated by variances $\sigma_j^2$, motivating top-variance selection.

\section{Implementation}
\label{sec:impl}

In this section, we present the kernel optimizations and system integration of \SystemName, detailing how the \splicegemm operator and TIM-aware sequence filtering are realized with negligible overhead over a native NVFP4 execution stack.

\subsection{The \splicegemm Kernel}
\label{subsec:splicegemm_kernel}

A naive realization of Equation~\ref{eq:splicegemm} that launches separate NVFP4 and BF16 GEMMs and merges their outputs in global memory incurs substantial kernel-launch overhead and memory-bandwidth waste. To optimize this, \splicegemm fuses the sparse BF16 compensation $X_S {W'_S}^\top$ into the NVFP4 execution pipeline.

\begin{figure}[!tb]
    \centering
    \includegraphics[width=\linewidth]{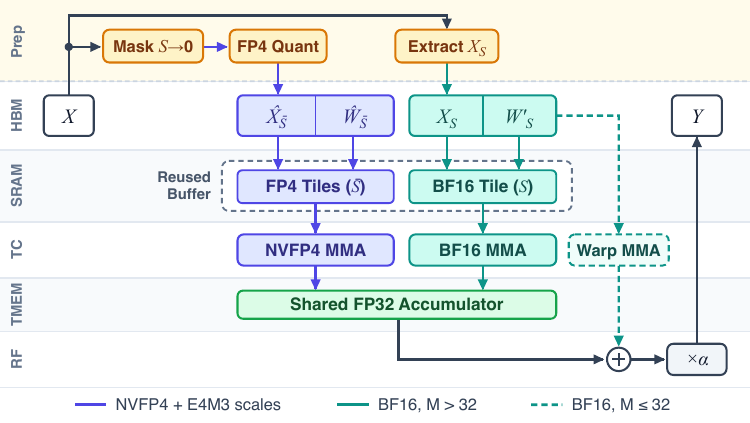}
    \caption{\splicegemm compute flow across the GPU and memory hierarchy. Operand preparation, Register Files, Tensor Cores, and Tensor Memory are denoted as \texttt{Prep}, \texttt{RF}, \texttt{TC}, and \texttt{TMEM}.}
    \label{fig:splicegemm-paths}
\end{figure}

\PHB{Data preparation.} As illustrated in Figure~\ref{fig:splicegemm-paths}, given an input activation $X \in \mathbb{R}^{M \times K}$ in HBM ($M$ denotes the number of input tokens), the preparation kernel first reads $X$ in a single pass using 256-bit vectorized loads. Each load instruction reads 16 contiguous BF16 elements per thread, matching one NVFP4 micro-block. Rather than branching per channel, each thread applies a precomputed register bitmask to zero out protected lanes in $S$ (\textbf{Mask} $S \to 0$) before quantizing the residual channels $\bar{S}$ into packed NVFP4 payloads $\widehat{X}_{\bar{S}}$ and block scales (\textbf{FP4 Quant}). Simultaneously, it writes the unmasked BF16 columns to a compact side buffer $X_S \in \mathbb{R}^{M \times |S|}$ (\textbf{Extract} $X_S$). The weight tensor $W$ is pre-split into a NVFP4 pack $\widehat{W}_{\bar{S}}$ and pre-a scaled BF16 pack ($W'_S = \operatorname{BF16}(W_S / \alpha) \in \mathbb{R}^{N \times |S|}$) after each policy update. This split is fixed during one rollout step.

\PHM{Dual-path execution.}
Because RL workloads span both compute-bound large batches (rollout prefill and training) and latency-sensitive small batches (autoregressive decoding), \splicegemm dispatches between two specialized compute paths according to the active token count $M$:
\begin{itemize}[topsep=0pt, partopsep=0pt, itemsep=0pt, parsep=0pt]
    \item \textbf{High-throughput prefill path ($M > 32$):} For large $M$, \splicegemm optimizes throughput via avoiding redundant memory allocations. Specifically, the kernel reuses the same SRAM buffer (\textbf{Reused Buffer}) to load the BF16 tile $(X_S, W'_S)$ after the NVFP4 MMA operation. Then, the BF16 MMA results are directly written into the same shared FP32 accumulator in TMEM. The combined result is then moved to RF, scaled by $\times \alpha$, and written to HBM as $Y$.
    \item \textbf{Low-latency decode path ($M \le 32$):} In the small-$M$ decode regime, TMA setup and shared-memory barriers for a narrow $M \times |S|$ tile incur disproportionate latency. While the main NVFP4 tiles execute via NVFP4 MMA, the compact side operands $(X_S, W'_S)$ bypass SRAM staging altogether and load directly from HBM into RF. Epilogue warps prefetch the $|S|$ protected weight columns into registers once per CTA and compute the correction via warp-level \texttt{mma.sync} instructions (\textbf{Warp MMA}), adding it to the TMEM main accumulator in RF right before the writeback.
\end{itemize}

\subsection{System Implementation}
\label{subsec:sys_impl}

We implement \SystemName atop veRL~\cite{sheng2024hybridflow}. Across both vLLM rollout and Megatron-LM training forward passes, linear projections are replaced with \splicegemm. To preserve full CUDA Graph capture in vLLM, execution plans cache compiled kernels and workspace descriptors strictly by $(M, N, K)$ shape without binding fixed tensor addresses. To track channels' variance, a lightweight forward hook uniformly subsamples token rows prior to masking to accumulate running channel moments in a contiguous GPU buffer. Before policy backpropagation, the trainer evaluates dual gates \texttt{G1} and \texttt{G2} and masks sequence advantages in-place, while backward passes run in BF16 with STE. After each optimizer step, \SystemName reselects protected channels according to the top-$s$ variance gathered in the previous step, with only $\sim 0.38\,\text{s}$ per update overhead. Packed weights $(\widehat{W}_{\bar{S}}, W'_S)$ are synchronized to vLLM via NCCL/CUDA IPC without re-quantization.

\section{Evaluation}
\label{sec:eval}
We evaluate whether \SystemName sustains FP4 training quality across engines and model sizes, how much its precision compensation costs, and how its two components contribute to stable training.

\begin{table*}[!t]
    \centering
    \caption{Qwen3-8B accuracy (\%) under FP4 inference (\%, Mean@16, temperature $T=0.6$, 8k response limit; mean $\pm$ standard error). QUADS and \SystemName retain their respective compensation at inference.}
    \label{table:main_results}
    {\scriptsize
        \setlength{\tabcolsep}{2.5pt}
        \begin{tabular}{l rrrrr rrrrr}
        \toprule
        \multirow{2}{*}{Method}
            & \multicolumn{5}{c}{\textbf{vLLM + Megatron-LM}}
            & \multicolumn{5}{c}{\textbf{vLLM + FSDP}} \\
        \cmidrule(lr){2-6} \cmidrule(lr){7-11}
            & AIME24 & AIME25 & AMC23 & MATH-500 & OCW
            & AIME24 & AIME25 & AMC23 & MATH-500 & OCW \\
        \midrule
        BF16   & 21.5$\pm$6.6 & 19.0$\pm$6.5 & 65.3$\pm$5.9 & 82.4$\pm$1.3 & 47.7$\pm$2.5 & \underline{23.1}$\pm$7.0 & 21.0$\pm$6.9 & 65.2$\pm$6.0 & 81.4$\pm$1.4 & 47.0$\pm$2.5 \\
        FP4    & 14.6$\pm$5.6 & 12.1$\pm$4.8 & 44.5$\pm$6.5 & 80.4$\pm$1.4 & 46.8$\pm$2.5 & 14.4$\pm$5.1 & 16.2$\pm$5.7 & 59.5$\pm$5.9 & 78.8$\pm$1.5 & 44.2$\pm$2.4 \\
        QaRL   & \underline{23.1}$\pm$6.4 & 19.4$\pm$6.3 & 67.8$\pm$6.0 & 83.5$\pm$1.3 & \textbf{49.3}$\pm$2.5 & 21.9$\pm$5.9 & \underline{21.3}$\pm$6.8 & \underline{66.7}$\pm$5.5 & \underline{81.8}$\pm$1.4 & 45.3$\pm$2.5 \\
        QuRL   & 6.7$\pm$3.4 & 2.7$\pm$2.0 & 30.9$\pm$5.9 & 68.1$\pm$1.7 & 42.2$\pm$2.4 & 11.7$\pm$4.6 & 10.8$\pm$4.3 & 54.1$\pm$6.1 & 77.3$\pm$1.6 & \underline{47.1}$\pm$2.6 \\
        QUADS  & 21.9$\pm$6.4 & \underline{21.9}$\pm$6.5 & \underline{70.6}$\pm$5.5 & \underline{83.7}$\pm$1.3 & 47.7$\pm$2.5 & 20.4$\pm$5.7 & 17.9$\pm$6.1 & 63.3$\pm$6.0 & 78.0$\pm$1.4 & 38.7$\pm$2.3\\
        \midrule
        \SystemName & \textbf{26.3}$\pm$6.6 & \textbf{23.3}$\pm$6.6 & \textbf{73.3}$\pm$5.4 & \textbf{85.9}$\pm$1.3 & \underline{48.4}$\pm$2.5 & \textbf{23.3}$\pm$6.5 & \textbf{21.3}$\pm$6.3 & \textbf{67.9}$\pm$5.5 & \textbf{82.5}$\pm$1.4 & \textbf{47.2}$\pm$2.5 \\
        \bottomrule
        \end{tabular}
    }
\end{table*}

\begin{figure*}[!t]
    \centering
    \includegraphics[width=0.9\linewidth]{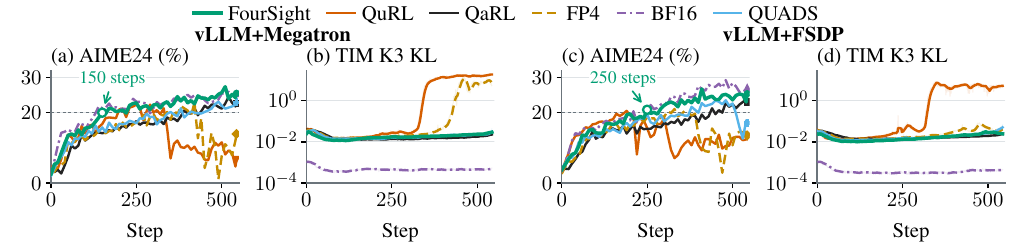}
    \vspace{-0.4cm}
    \caption{Qwen3-8B training with DAPO: AIME24 validation accuracy and training--inference mismatch on trained vLLM + Megatron-LM (a,b) and vLLM + FSDP (c,d). BF16 validation uses BF16 inference; other methods use FP4.}
    \label{fig:main_new_stack}
    \vspace{-0.4cm}
\end{figure*}

\subsection{Experimental Setup}
\PHM{Platform and workloads.} We use four nodes with eight NVFP4-capable GPUs each. Training uses Megatron-LM 0.19.2 or Fully Sharded Data Parallel (FSDP) with PyTorch 2.9.1; rollout uses vLLM 0.26.0. Our main workload trains Qwen3-8B~\cite{yang2025qwen3} on DAPO-Math-17k using DAPO algorithm~\cite{yu2025neurips-dapo}, with an 8K response limit. We also train Qwen3-1.7B on GSM8K using GRPO algorithm~\cite{shao2024deepseekmath} and a 512-token output limit. Performance experiments cover Qwen3-8B, Qwen3-32B, and Nemotron3-Super-120B-A12B~\cite{Chandiramani2026Nemotron3S} on 8, 16, and 32 GPUs, respectively.

\PHM{Methods.} We compare with BF16 training and rollout, native FP4 rollout with QAT, and three recent quantized RL methods: Quantization-error Alignment across Dual Sides (QUADS)~\cite{zhuge2026quads}, Quantized Rollout (QuRL)~\cite{li2026qurl}, and Rollout-Aligned Quantization-Aware RL (QaRL)~\cite{gu2026qarl}. All methods enable TIS~\cite{yao2025rollout} according to their default recipe. \SystemName combines sequence filtering (SF) with precision compensation (PC) for 64 input channels per projection. Filtering uses $\tau=10^{-4},K_1=0.01\times\text{output\_len}, K_2=5$ and channel reselection is performed every step according to the previously accumulated variance.

\PHM{Metrics.} Training curves use each method's rollout precision. Final checkpoints are evaluated with FP4 inference to reflect deployment requirements, including those trained in BF16. Each method's compensation is retained. For Qwen3-8B, we report Mean@16 on AIME24~\cite{aime2024}, AIME25~\cite{aime2025}, AMC23~\cite{amc23}, MATH-500~\cite{lightman2023let}, and OCW~\cite{lewkowycz2022solving}. For Qwen3-1.7B, we report Mean@16 on GSM8K~\cite{cobbe2021gsm8k}, SVAMP~\cite{patel2021nlp}, MATH-500~\cite{lightman2023let}, and GSM1K~\cite{zhang2024careful}. We measure TIM using sampled estimates of K3 KL divergence between the two policies.

\subsection{Training Quality}
\PHB{DAPO on Qwen3-8B.} Figure~\ref{fig:main_new_stack} compares training curves and the corresponding TIM on both engine combinations. \SystemName steadily improves validation accuracy and remains close to BF16 throughout training. Pure FP4 and QuRL make early gains but deteriorate later; QuRL's rapid initial progress gives way to a sharp increase in mismatch. QaRL remains stable on both engines, while QUADS loses accuracy late in FSDP training. We also measure learning speed by the first validation step reaching 20\% AIME24 accuracy. \SystemName reaches this target in 150 steps on Megatron and 250 steps on FSDP, Compared with QaRL and QUADS, this improves step efficiency by 2.60$\times$ and 2.33$\times$ on Megatron, and 1.76$\times$ and 1.52$\times$ on FSDP, respectively. 

Table~\ref{table:main_results} shows that fast, stable learning translates into strong FP4 deployment quality. \SystemName achieves the highest accuracy on four of five benchmarks with Megatron and leads all five with FSDP. On Megatron, it improves AIME24 from 23.1\% for the strongest baseline to 26.3\%, and AMC23 from 70.6\% to 73.3\%. On FSDP, it improves every benchmark over the BF16-trained model and matches or surpasses all quantized baselines, combining faster learning with consistent gains across tasks.

The engine combination matters because it changes the numerical errors encountered during policy updates: Megatron executes native FP4 forward operations, whereas FSDP emulates quantization in higher precision via QAT. The resulting failure patterns also differ. Pure FP4 under Megatron exhibits a large late-stage KL increase; under FSDP, substantial accuracy loss occurs even while average KL remains small. \SystemName addresses both rollout quality and harmful trajectories, sustaining training across these different error patterns. We isolate the two contributions below.

\begin{figure}[!tb]
    \centering
    \includegraphics[width=\linewidth]{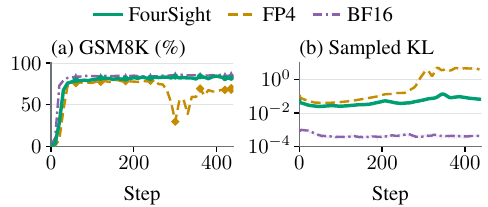}
    \vspace{-0.4cm}
    \caption{Qwen3-1.7B training with GRPO: \SystemName sustains close-to-BF16 accuracy and controls TIM as pure FP4 degrades.}
    \label{fig:gsm8k_val_kl}
    \vspace{-0.4cm}
\end{figure}

\PHM{GRPO on Qwen3-1.7B.} \SystemName generalizes effectively to smaller models and distinct algorithms like GRPO (Figure~\ref{fig:gsm8k_val_kl}). While pure FP4 suffers severe accuracy degradation as policy mismatch accumulates, \SystemName stably maintains $\sim$83\% validation accuracy, closely matching the BF16 baseline. In native FP4 deployment, \SystemName outperforms all baselines across four benchmarks (Table~\ref{tab:gsm8k-1p7b-eval}), exceeding the BF16 baseline by 12.7--16.4\,pp. This superiority raises because standard BF16-trained models suffer heavy accuracy loss when subjected to FP4 post-training quantization (PTQ), while \SystemName inherently immunizes the policy against low-bit errors during training, preserving reasoning capabilities directly in the target serving precision.

\begin{table}[!tb]
    \centering
    \caption{Qwen3-1.7B FP4 inference accuracy (\%, Mean@16, temperature $T=0.6$, 4k response limit; mean $\pm$ standard error).}
    \label{tab:gsm8k-1p7b-eval}
    \scriptsize
    \setlength{\tabcolsep}{5pt}
    \begin{tabular}{lrrrr}
    \toprule
    Method & GSM8K & SVAMP & MATH-500 & GSM1K \\
    \midrule
    BF16 & 67.30$\pm$0.97 & 74.54$\pm$0.96 & 37.82$\pm$1.63 & 59.13$\pm$1.07 \\
    FP4 & 45.86$\pm$1.10 & 49.67$\pm$1.28 & 16.25$\pm$1.24 & 39.38$\pm$1.13 \\
    \SystemName & \textbf{80.65}$\pm$0.83 & \textbf{87.26}$\pm$0.81 & \textbf{51.34}$\pm$1.77 & \textbf{75.53}$\pm$0.95 \\
    \bottomrule
    \end{tabular}
    \vspace{-0.4cm}
\end{table}

\subsection{Performance and Overhead}
\begin{figure}[!tb]
    \centering
    \includegraphics[width=\linewidth]{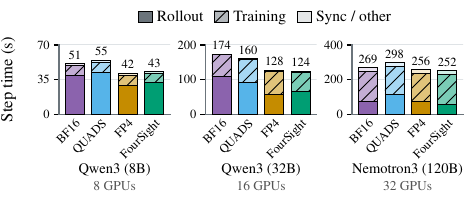}
    \vspace{-0.4cm}
    \caption{Step time and breakdown at matched response lengths. \SystemName introduces less than 2.3\% overhead to pure FP4.}
    \label{fig:rl_step_time}
\end{figure}

\PHB{End-to-end performance.} To ensure fair comparisons, we evaluate all methods on identical training trajectories with fixed prompt and response length distribution. Figure~\ref{fig:rl_step_time} reports the average per-step time and breakdown after warmup, where \SystemName consistently retains near-FP4 step time. \SystemName completes a step in 43\,s on Qwen3-8B and 124\,s on Qwen3-32B, yielding 1.18$\times$ and 1.40$\times$ speedups over the BF16 baseline, and 1.28$\times$ and 1.29$\times$ over QUADS. Notably, \SystemName introduces negligible overhead compared to uncompensated pure FP4: it is only 2.3\% slower on 8B and even 3.1\% faster on 32B. This efficiency is driven by our \splicegemm kernel, which not only hides compensation latency but also outperforms native FlashInfer under memory-bound decode regimes on Qwen3-32B. For Nemotron3, \SystemName takes 252\,s per step, 1.56\% below pure FP4 and 15.4\% below QUADS (1.18$\times$ faster). Since training stage dominates this workload, faster low-precision forward operations during rollout have a smaller gain on total time.

\begin{figure}[!tb]
    \centering
    \includegraphics[width=\linewidth]{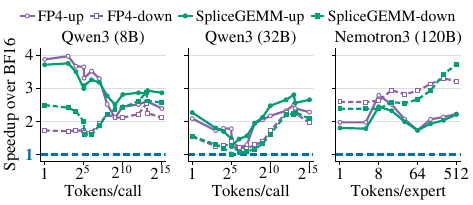}
    \vspace{-0.4cm}
    \caption{Kernel performance measured by speedup over BF16 with 64-channel compensation. ``Up'' and ``down'' denote the two feed-forward projections; the blue line marks BF16 speed.}
    \label{fig:splicegemm_performance}
    \vspace{-0.4cm}
\end{figure}

\PHM{Kernel performance.} Figure~\ref{fig:splicegemm_performance} compares complete projection operations, including activation quantization, across token counts. \splicegemm reaches up to 3.7$\times$ and 2.8$\times$ BF16 speed on Qwen3-8B and 32B, respectively, and broadly tracks pure FP4. For Nemotron3, \splicegemm achieves up to 2.4$\times$ and 3.7$\times$ speedup over the vLLM BF16 baseline for Up and Down, respectively. The optimized Down stage also outperforms pure FP4 by up to 1.16$\times$. Fusing the correction into FP4 kernels retains the throughput benefits of low-precision execution while protecting selected channels.

\subsection{Mechanism Ablations}

\begin{figure*}[!t]
    \centering
    \includegraphics[width=0.9\linewidth]{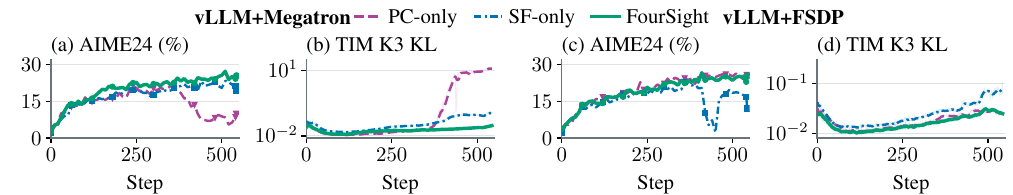}
    \vspace{-0.4cm}
    \caption{Ablation study of sequence filtering (SF), precision compensation (PC), and their combination across training engines.}
    \label{fig:ablation_comparison}
    \vspace{-0.4cm}
\end{figure*}

\PHB{Filtering and compensation.} Figure~\ref{fig:ablation_comparison} isolates the contributions of Precision Compensation (PC) and Selective Filtering (SF). On Megatron, PC alone triggers eventual policy collapse, whereas SF alone stabilizes training but plateaus at a suboptimal accuracy. Unifying both interventions boosts final AIME24 accuracy from 20.8\% (SF-only) to 26.3\% (Table~\ref{tab:ablation}). This synergy arises because SF purges severe out-of-distribution trajectories to shield policy updates from destabilization, while PC elevates the baseline generation quality of the remaining trajectories to provide richer learning signals.

Under FSDP, PC alone recovers most accuracy gains, achieving the highest scores on AIME25 and MATH-500. Yet, adding SF improves OCW accuracy over PC-only, while adding PC prevents the late-stage degradation observed with SF alone. These behaviors reveal that while the dominance of each component shifts with engine-specific numerical discrepancies, their joint application is necessary to guarantee robust, high-quality RL across divergent execution frameworks.

\begin{table}[!tb]
    \centering
    \caption{Final FP4 accuracy (\%) with sequence filtering (SF), precision compensation (PC), or both. Evaluation follows Table~\ref{table:main_results}.}
    \label{tab:ablation}
    {\scriptsize
        \setlength{\tabcolsep}{5pt}
        \begin{tabular}{lrrrrr}
        \toprule
        Method & AIME24 & AIME25 & AMC23 & MATH-500 & OCW \\
        \cmidrule(lr){2-6}
        & \multicolumn{5}{c}{\textbf{vLLM + Megatron-LM}} \\
        \cmidrule(lr){2-6}
        SF-only & \underline{20.8}$\pm$6.0 & \underline{19.2}$\pm$6.0 & \underline{69.7}$\pm$5.2 & \underline{83.4}$\pm$1.3 & \underline{48.6}$\pm$2.5 \\
        PC-only & 10.0$\pm$4.3 & 4.4$\pm$2.6 & 40.6$\pm$5.9 & 69.1$\pm$1.7 & 41.2$\pm$2.4 \\
        \SystemName & \textbf{26.3}$\pm$6.6 & \textbf{23.3}$\pm$6.6 & \textbf{73.3}$\pm$5.4 & \textbf{85.9}$\pm$1.3 & 48.4$\pm$2.5 \\
        \midrule
        & \multicolumn{5}{c}{\textbf{vLLM + FSDP}} \\
        \cmidrule(lr){2-6}
        SF-only & 12.5$\pm$4.7 & 14.8$\pm$5.4 & 50.5$\pm$6.6 & 81.0$\pm$1.4 & \textbf{47.4}$\pm$2.6 \\
        PC-only & \underline{23.3}$\pm$6.1 & \textbf{22.1}$\pm$6.7 & \textbf{67.9}$\pm$5.5 & \textbf{84.7}$\pm$1.3 & 44.0$\pm$2.5 \\
        \SystemName & \textbf{23.3}$\pm$6.5 & \underline{21.3}$\pm$6.3 & \underline{67.9}$\pm$5.5 & \underline{82.5}$\pm$1.4 & \underline{47.2}$\pm$2.5 \\
        \bottomrule
        \end{tabular}
    }
\end{table}

\begin{figure}[!tb]
    \centering
    \includegraphics[width=\linewidth]{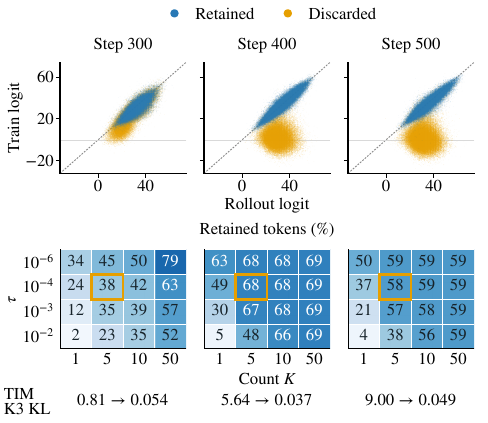}
    \vspace{-0.6cm}
    \caption{Sensitivity of the low-probability token criterion. Top: retained and discarded tokens. Bottom: token retention versus probability threshold $\tau$ and count $K$; gold marks the default. The footer reports TIM KL before and after filtering.}
    \label{fig:zeroout_sensitivity}
\end{figure}

\PHB{Sequence filtering sensitivity.} Figure~\ref{fig:zeroout_sensitivity} analyzes the tail-token filter across three Qwen3-8B training snapshots on DAPO-Math-17k. At our default threshold ($\tau=10^{-4}$ and $K=5$) the filter eliminates severely divergent off-diagonal logit clusters while retaining 38--68\% of tokens, collapsing token-level KL divergence from 0.81--9.00 to 0.037--0.054. This aligns the training trajectory tightly within policy consensus. Stricter filtering discards excessive training data, while larger looser guards ($K>5$) add little retained data at later training steps with increased TIM KL. This validates our configuration as an effective trade-off between mismatch suppression and data efficiency.

\begin{figure}[!tb]
    \centering
    \includegraphics[width=\linewidth]{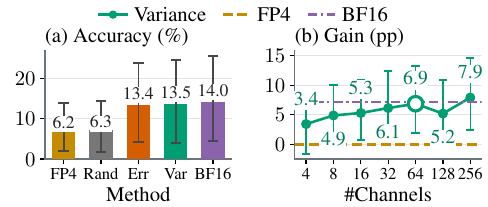}
    \vspace{-0.4cm}
    \caption{Frozen-checkpoint AIME24 accuracy with random (Rand), error-based (Err), or variance-based (Var) channel selection; right: gain in percentage points (pp) versus channel budget. Bars show 95\% confidence intervals over 30 samples.}
    \label{fig:channel_protection_s525}
    \vspace{-0.5cm}
\end{figure}

\PHM{Precision compensation sensitivity.} We benchmark a frozen, degraded Qwen3-8B checkpoint across varying compensation schemes (Figure~\ref{fig:channel_protection_s525}). Protecting merely 64 variance-selected channels elevates AIME24 accuracy to 13.5\% (nearing BF16's 14.0\%), whereas random channel protection fails completely, hovering near uncompensated FP4. This demonstrates the importance of channel selection, where minimal, targeted precision safeguards can recover generation quality. Scaling to 256 channels yields a marginal $\sim$1\,pp gain despite a 4$\times$ increase in protected state. This confirms that 64 channels captures the vast majority of accuracy gains at minimal hardware overhead.

\section{Related Work}
\label{sec:related}

\subsection{Quantized Inference and Learning}
\label{subsec:related_quant}
To improve low-bit severing quality, post-training quantization (PTQ) and inference-time enhancements mitigate activation outliers via channel smoothing~\cite{xiao2023smoothquant, lin2024awq}, orthogonal rotations~\cite{ashkboos2024quarot}, residual channel augmentation~\cite{meng2026arcquant}, or mixed-precision decomposition~\cite{dettmers2022gpt3, zhao2024atom}. Beyond static inference, quantization-aware training (QAT)~\cite{jacob2017quantization, moonshot2025kimi} and low-bit pre-training~\cite{micikevicius2022fp8, wang2025fp4, nvidia2024blackwell} adapt parameters to low-precision arithmetic during single-engine teacher-forcing updates. Directly applying them to NVFP4 RL is insufficient: policy updates amplify quantization sensitivity over long rollouts, while online rotations or unaligned compensation branches add latency and cross-engine drift. \SystemName bounds NVFP4 block-scaling error via top-variance channel selection and fused \splicegemm kernel with specialized prefill and decode paths.

\subsection{Remedies for Training--Inference Mismatch}
\label{subsec:related_tim}
Decoupled rollout and training engines induce Training--Inference Mismatch (TIM)~\cite{yao2025offpolicy, yao2025rollout}. System unification like veXact~\cite{zhong2026vexact} eliminates TIM by sharing one engine, but forfeits stage-specific optimizations. Algorithmic remedies adjust updates via TIS~\cite{yao2025rollout}, Score Centering~\cite{marek2026score}, token-level masking~\cite{lingteam2025stepevolves}, sequence-level TRM~\cite{li2026trm}, or TRIAGE~\cite{li2026triage}. For quantized RL, QuRL~\cite{li2026qurl} pairs adaptive ratio clipping with invariant weight scaling in 8-bit rollouts; QaRL~\cite{gu2026qarl} combines rollout-aligned learner forwards with sequence-level trust-band clipping; and HiFloat4~\cite{mak2026hifloat4} and QUADS~\cite{zhuge2026quads} compensate FP4 rollout errors via auxiliary residual GEMMs. Under NVFP4, token-level corrections still leak corrupted advantages, whereas sequence filtering cannot restore degraded rollout quality. Meanwhile, existing compensations incur severe overhead and leave outliers unfiltered. \SystemName co-designs dual-gate sequence filtering with fused \splicegemm compensation for both stability and near-native FP4 speed.

\section{Conclusion}
\label{sec:conclusion}

We presented \SystemName, an algorithm--system co-design that enables stable and efficient NVFP4 RL post-training. 
Through fine-grained diagnostics, we revealed that low-bit RL failures are sparse rather than uniform, motivating two complementary interventions: a dual-gate sequence filtering to isolate drifted trajectories from policy updates, and sparse precision compensation that protects only top-variance channels in BF16. Powered by our fused \splicegemm kernel, \SystemName sustains near-BF16 training stability across scales and achieves state-of-the-art FP4 deployment accuracy while remaining within $2.3\%$ of native FP4 step time up to 120B-scale models. These results demonstrate that algorithm-system co-design can overcome the longstanding trade-off between low-bit efficiency and training stability, providing a practical foundation for efficient RL post-training and deployment at scale.

\clearpage

\bibliography{refs}
\bibliographystyle{mlsys2025}

\clearpage

\section{Appendix}
\label{append: theory_proof}

\subsection{Degenerate words}

\subsection{Experiment Settings}
\label{sec:experiment_settings}
As discussed in Section~\ref{sec:eval}, Table~\ref{tab:grpo_dapo_recipe} provides a detailed summary of the experimental settings for the GRPO and DAPO recipes. For DAPO, we also enable group filtering based on accuracy and configure the overlong buffer with a length of 512 and a penalty factor of 1.0 to penalize the overlong sequences. For QaRL~\cite{gu2026qarl} and QuRL~\cite{li2026qurl} reproduction, all algorithm-specific hyperparameters are kept at the values reported in the original papers.

\begin{table}[h]
  \centering
  \scriptsize
  \caption{Experimental settings for the GRPO and DAPO recipes.}
  \label{tab:grpo_dapo_recipe}
  \setlength{\tabcolsep}{5pt}
  \begin{tabular}{lcc}
    \toprule
    \textbf{Parameter Name} & \textbf{GRPO (1.7B)} & \textbf{DAPO (8B)} \\
    \midrule
    \texttt{trainer.nnodes} & 1 & 1 \\
    \texttt{trainer.n\_gpu\_per\_node} & 8 & 8 \\
    \texttt{data.train\_batch\_size} & 256 & 32 \\
    \texttt{data.max\_prompt\_length} & 1024 & 1024 \\
    \texttt{data.max\_response\_length} & 512 & 8192 \\
    \texttt{rollout.n} & 4 & 8 \\
    \texttt{rollout.temperature} & 1.0 & 1.0 \\
    \texttt{rollout.top\_p} & 1.0 & 1.0 \\
    \texttt{val\_kwargs.temperature} & 0 & 0.6 \\
    \texttt{actor.ppo\_mini\_batch\_size} & 256 & 32 \\
    \texttt{optimizer} & AdamW & AdamW \\
    \texttt{optim.lr} & $1\times10^{-6}$ & $1.5\times10^{-6}$ \\
    \texttt{optim.weight\_decay} & 0.01 & 0.1 \\
    \texttt{rollout\_is} & token & token \\
    \texttt{rollout\_is\_threshold} & 2.0 & 2.0 \\
    \bottomrule
  \end{tabular}
\end{table}

\subsection{Error Bounds of Variance-based Compensation}
\label{sec:variance_proof}

\subsubsection{Notation and preliminaries}
\label{sec:variance64_theory}

\PHB{The linear layer and the protected set under quantization.}
\label{sec:notation_layer}
We consider a linear layer
\[
y = Wx,
\qquad
W \in \mathbb{R}^{N \times K},
\qquad
x \in \mathbb{R}^{K},
\]
and study the expected squared output reconstruction error relative to
the BF16 weights and activations.

The implementation protects a fixed set \(S \subseteq \{1,\ldots,K\}\)
of input channels, with \(\lvert S\rvert=s\) using a BF16 side branch.
The remaining channels are denoted by$R = \{1,\ldots,K\}\setminus S$. 

For notational convenience, \(x_R\) and \(W_R\) denote tensors embedded
in the original \(K\)-dimensional coordinate system, with the
coordinates in \(S\) set to zero. Similarly, \(x_S\) and \(W_S\) retain
only the protected coordinates. Thus,
\[
x=x_R+x_S,
\qquad
W=W_R+W_S.
\]
Since \(W_R\) vanishes on the columns in \(S\) and \(x_R\) vanishes on
the coordinates in \(S\), and symmetrically for \(W_S,x_S\) on \(R\),
we have \(W_Rx_S=0\) and \(W_Sx_R=0\). Hence
\begin{equation}
y = W_Rx_R + W_Sx_S.
\label{eq:output_decomposition}
\end{equation}

After the selected coordinates are zeroed, the implementation
recomputes the local NVFP4 block scales. Write 
\(\mathcal{Q}_A^S\) and \(\mathcal{Q}_W^S\) for the activation and weight
quantizers
. We write
\(\mathcal{Q}_A\) for the unprotected baseline activation quantizer,
which corresponds to \(S=\varnothing\). Ignoring the finite-precision
error of the BF16 side branch for the moment, the idealized protected
output is
\begin{equation}
\widehat y_S
=
\mathcal{Q}_W^S(W_R)\,
\mathcal{Q}_A^S(x_R)
+
W_Sx_S.
\label{eq:protected_output}
\end{equation}

\PHM{The NVFP4 format.}
\label{sec:notation_nvfp4}
The finite E2M1 value set is
\[
\mathcal{C}
=
\{0,\pm 0.5,\pm 1,\pm 1.5,\pm 2,\pm 3,\pm 4,\pm 6\}.
\]
The largest gap between adjacent nonnegative values of \(\mathcal{C}\)
is \(2\), and its largest magnitude is \(6\).

Consider one activation block \(b\) (of width \(B\) along \(K\)) after
the protected coordinates have been zeroed, and define
\[
a_b(S)
=
\max_{j\in b\cap R}\lvert x_j\rvert.
\]
Let \(g_A>0\) be the frozen activation encode gain. The local physical
scale with the E4M3 block scale factor is
\begin{equation}
\widetilde s_b
=
\frac{
\operatorname{round}_{\mathrm{E4M3}}
\left(g_Aa_b(S)/6\right)
}{g_A}.
\label{eq:nvfp4_scale}
\end{equation}
For E4M3 normal numbers, the unit roundoff is
\(u_{\mathrm{E4M3}}=1/16\). Weight blocks are defined analogously along
each output row, with the corresponding weight gain.

\PHM{Activation moments and weight perturbations.}
\label{sec:notation_moments}

Define the first moment $\mu = \mathbb{E}[x]$, and the second moment $H$ by 
\[
\Sigma
=
\mathbb{E}\!\left[(x-\mu)(x-\mu)^{\mathsf T}\right],
\quad
H
=
\mathbb{E}[xx^{\mathsf T}]
=
\Sigma+\mu\mu^{\mathsf T}.
\]
Write the variance and second moment for channel $j$ as
\[
v_j=\Sigma_{jj},
\ \ 
m_j=\mathbb{E}[x_j^2]=v_j+\mu_j^2,
\]
with  \(v_{(1)}\geq\cdots\geq v_{(K)}\) and
\(m_{(1)}\geq\cdots\geq m_{(K)}\).

Let $D = \widehat W-W$
be a fixed weight perturbation, and let \(\delta w_j\) denote column
\(j\) of \(D\). Let \(w_j\) denote column \(j\) of \(W\), and write
\(\widehat w_j=w_j+\delta w_j\).

\begin{lemma}[Exact output error induced by a weight perturbation]
\label{lem:weight_quadratic}
The expected squared output error is
\begin{align}
\mathcal{E}_W
&=
\mathbb{E}\!\left[\lVert Dx\rVert_2^2\right]
\nonumber\\
&=
\operatorname{tr}\!\left(DHD^{\mathsf T}\right)
\nonumber\\
&=
\operatorname{tr}\!\left(D\Sigma D^{\mathsf T}\right)
+
\lVert D\mu\rVert_2^2.
\label{eq:exact_weight_error}
\end{align}
\end{lemma}

\begin{proof}
Using
\(
\lVert Dx\rVert_2^2
=
x^{\mathsf T}D^{\mathsf T}Dx
\)
and the cyclic property of the trace,
\[
\mathbb{E}\!\left[\lVert Dx\rVert_2^2\right]
=
\operatorname{tr}
\left(
D\,\mathbb{E}[xx^{\mathsf T}]\,D^{\mathsf T}
\right)
=
\operatorname{tr}(DHD^{\mathsf T}).
\]
Substituting \(H=\Sigma+\mu\mu^{\mathsf T}\) gives
\[
\operatorname{tr}(D\mu\mu^{\mathsf T}D^{\mathsf T})
=
\lVert D\mu\rVert_2^2,
\]
which proves the result.
\end{proof}

\subsubsection{Proof Structure}
\label{sec:proof_structure}

The justification of variance-based protection rests on two pillars, which are then joined into a single bound. The first pillar is that channels with high second moment amplify weight error, motivating us to select according to \emph{which} quantity. The second pillar is that controlling residual block magnitude can control NVFP4 error. It explains \emph{why} removing channels helps beyond the removed channel itself. 

\PHM{Pillar I: Activation energy amplifies weight error.} With Lemma~\ref{lem:weight_quadratic}, 
the expected output error with weight perturbation \(D\) can be expressed as 
\(\operatorname{tr}(D\Sigma D^{\mathsf T})+\lVert D\mu\rVert_2^2\). According to this formula, we can prove that channel \(j\) contributes \(m_j\lVert\delta w_j\rVert_2^2\) error 
(Equation~\ref{eq:direct_weight_gain}) under a diagonal approximation. Thus we should consider protecting channels with large 
second moment \(m_j\). 

\PHM{Pillar II: NVFP4 error is controlled by residual block magnitude.}
Within each block, the element error is at most \((17/96)a_b(S)\)
(Lemma~\ref{lem:fp4_element_bound}). Therefore,  removing a dominant channel lowers can shrink the error of its block neighbors. 
Lifting from blocks to norms gives the \(\sqrt{B}\)-scaled bounds of
Lemma~\ref{lem:block_norm_bound}. 

\paragraph{Bridge: A separable upper bound.}
Combining the error decomposition in Pillar I with the bounds in Pillar II yields that the quantization error with protection subset S can be bounded by 
\(\mathbb{E}\lVert\widehat y_S-y\rVert_2^2
\le C\lVert W\rVert_F^2\sum_{j\notin S}m_j\)
(Proposition~\ref{prop:residual_energy_bound}). 
Because this relaxed bound $C\lVert W\rVert_F^2\sum_{j\notin S}m_j$ is separable, it can be  minimized by  selecting the s channels with the largest second moments (Proposition~\ref{prop:top_second_moment}). Under the mean-gap condition, it is minimized by Top-\(s\) variance (Corollary~\ref{cor:variance_ranking}). That is our selection policy. Main theorem~\ref{thm:variance_protection} collects these statements into the target result: protecting the Top-\(s\) variance channels bounds
the quantization error.

\subsubsection{Proof of the Error Bound}
\label{sec:proof}

\PHB{Pillar I: Activation energy amplifies weight error.}
\label{sec:pillar1}
We first keep the activation in high precision.
For each output row, the Hessian of the squared reconstruction loss
with respect to that row of the weight matrix is \(2H\). If the
objective measures only the variance of the output perturbation around
its mean, the relevant quadratic form uses \(\Sigma\) instead.

Let \(v_j=\Sigma_{jj}\) be the activation variance and let
\(\delta w_j\) be column \(j\) of \(D\). Under the diagonal
approximation
\[
\Sigma \approx \operatorname{diag}(v_1,\ldots,v_K),
\]
Equation~\ref{eq:exact_weight_error} becomes
\begin{equation}
\mathcal{E}_W
\approx
\sum_{j=1}^{K}
v_j\lVert\delta w_j\rVert_2^2
+
\lVert D\mu\rVert_2^2.
\label{eq:diagonal_weight_error}
\end{equation}

If \(H\) is diagonal and the quantization errors of all other columns
are held fixed, eliminating the weight error of channel \(j\) reduces
the total MSE by exactly
\begin{equation}
g_j^{(W)}
=
m_j\lVert\delta w_j\rVert_2^2,
\qquad
m_j
=
\mathbb{E}[x_j^2]
=
v_j+\mu_j^2.
\label{eq:direct_weight_gain}
\end{equation}
For a centered objective with \(\mu=0\), this reduces to
\[
g_j^{(W)}
=
v_j\lVert\delta w_j\rVert_2^2.
\]

Equation~\ref{eq:direct_weight_gain} requires the diagonal
approximation and the absence of mean-induced cross-column
cancellation. Merely holding the other quantized columns fixed is not
sufficient when \(H\) is non-diagonal.

\PHM{Pillar II: NVFP4 error is controlled by residual block
magnitude.}
\label{sec:pillar2}
We first present a simple example for \emph{Neighbor effect of protection.} The ideal-scale quantization of $(48,1,\ldots,1)$
uses scale \(8\). The value \(48\) is exactly representable, while all
entries equal to \(1\) round to zero. If the channel containing \(48\)
is protected, the residual block has scale \(1/6\), and the entries
equal to \(1\) become exactly representable. Thus, protecting a
dominant channel can improve its block neighbors even if the dominant
activation itself had zero activation-quantization error. Next we will prove the bound. 

\emph{\underline{Per-element bound with E4M3 rounding.}}
We first account for E4M3 scale rounding per-element.

\begin{lemma}[Per-element NVFP4 error bound]
\label{lem:fp4_element_bound}
Assume that the post-masking local scale is an E4M3 normal number and
does not saturate. Let
$
\widetilde s_b
=
\frac{a_b(S)}{6}(1+\varepsilon_b),
\ 
\lvert\varepsilon_b\rvert\leq u_{\mathrm{E4M3}} =\frac{1}{16}$  
Then every residual coordinate in the block satisfies
\begin{equation}
\left|
\mathcal{Q}_A^S(x_j)-x_j
\right|
\leq
c_A a_b(S),
\qquad
c_A=\frac{17}{96}.
\label{eq:fp4_element_bound}
\end{equation}
The same statement applies to each weight block, with corresponding constant \(c_W\).
\end{lemma}

\begin{proof}
For a value inside the representable E2M1 range, the nearest-rounding
error is at most \(\widetilde s_b\), because the largest adjacent E2M1
gap is \(2\widetilde s_b\). If the rounded scale is smaller than the
ideal scale, the maximum element may exceed the endpoint
\(6\widetilde s_b\). The resulting clipping error is at most
\[
\left(a_b(S)-6\widetilde s_b\right)_+.
\]
Hence,
\[
\left|
\mathcal{Q}_A^S(x_j)-x_j
\right|
\leq
\max
\left\{
\widetilde s_b,
\left(a_b(S)-6\widetilde s_b\right)_+
\right\}.
\]
The first term satisfies
\[
\widetilde s_b
\leq
\frac{1+1/16}{6}a_b(S)
=
\frac{17}{96}a_b(S).
\]
The second term is bounded by
\[
\left(a_b(S)-6\widetilde s_b\right)_+
\leq
\frac{1}{16}a_b(S)
\leq
\frac{17}{96}a_b(S).
\]
This proves Equation~\ref{eq:fp4_element_bound}.
\end{proof}

The normal-scale assumption must hold after masking. Removing the
largest channel can reduce the residual scale into the E4M3 subnormal
range, where a uniform relative error bound of \(1/16\) no longer
holds. 
Scale underflow therefore requires either a separate absolute
error term or an empirical check.

\emph{\underline{Lifting from blocks to norms.}}
Define the selection-dependent quantization
errors
\[
\Delta x_S
=
\mathcal{Q}_A^S(x_R)-x_R,
\qquad
D_S
=
\mathcal{Q}_W^S(W_R)-W_R.
\]
Then for the unprotected baseline, the per-channel activation error is $e_j
=
\mathcal{Q}_A(x_j)-x_j.$

\begin{lemma}[Blockwise norm bounds]
\label{lem:block_norm_bound}
Under the assumptions of Lemma~\ref{lem:fp4_element_bound}, for block
width \(B\),
\begin{align}
\lVert\Delta x_S\rVert_2
&\leq
\sqrt{B}\,c_A\lVert x_R\rVert_2,
\label{eq:activation_norm_bound}
\\
\lVert D_S\rVert_F
&\leq
\sqrt{B}\,c_W\lVert W_R\rVert_F.
\label{eq:weight_norm_bound}
\end{align}
\end{lemma}

\begin{proof}
For one activation block \(b\), let
\(
n_b=\lvert b\cap R\rvert\leq B
\).
Lemma~\ref{lem:fp4_element_bound} gives
\[
\sum_{j\in b\cap R}
\left|
\mathcal{Q}_A^S(x_j)-x_j
\right|^2
\leq
n_bc_A^2a_b(S)^2.
\]
Since $a_b(S)^2
\leq
\sum_{j\in b\cap R}x_j^2$, 
we obtain
\[
\sum_{j\in b\cap R}
\left|
\mathcal{Q}_A^S(x_j)-x_j
\right|^2
\leq
Bc_A^2
\sum_{j\in b\cap R}x_j^2.
\]
Summing over blocks proves
Equation~\ref{eq:activation_norm_bound}. Applying the same argument
to every output-row weight block proves
Equation~\ref{eq:weight_norm_bound}.
\end{proof}

\PHM{Bridge: A separable upper bound.}
\label{sec:bridge}
First, we present a key proposition by combining the two pillars, inducing a quantization error upper bound expressed in terms of the residual second moments.   

\begin{proposition}[Residual-energy upper bound]
\label{prop:residual_energy_bound}
Under the assumptions above, the idealized protected output in
Equation~\ref{eq:protected_output} satisfies
\begin{align}
\mathbb{E}
\left[
\lVert\widehat y_S-y\rVert_2^2
\right]
&\leq
C
\lVert W_R\rVert_F^2
\sum_{j\notin S}m_j
\label{eq:tighter_upper_bound}
\\
&\leq
C
\lVert W\rVert_F^2
\sum_{j\notin S}m_j,
\label{eq:relaxed_upper_bound}
\end{align}
where
\begin{equation}
C
=
\left[
\sqrt{B}\,c_W
+
\sqrt{B}\,c_A
\left(1+\sqrt{B}\,c_W\right)
\right]^2.
\label{eq:upper_bound_constant}
\end{equation}
For \(B=16\),
\[
C
=
\left[
4c_W+4c_A(1+4c_W)
\right]^2.
\]
With \(c_A=c_W=17/96\), this gives \(C\approx 3.68\).
\end{proposition}

\begin{proof}
By Equations~\ref{eq:output_decomposition}
and~\ref{eq:protected_output}, the term \(W_Sx_S\) cancels.
Expanding the main-branch error gives
\begin{align}
\widehat y_S-y
&=
(W_R+D_S)(x_R+\Delta x_S)-W_Rx_R
\nonumber\\
&=
D_Sx_R+(W_R+D_S)\Delta x_S.
\end{align}
Using the triangle inequality and
\(
\lVert A\rVert_{\mathrm{op}}\leq\lVert A\rVert_F
\),
\begin{align}
\lVert\widehat y_S-y\rVert_2
&\leq
\lVert D_S\rVert_F\lVert x_R\rVert_2
+
\left(
\lVert W_R\rVert_F+\lVert D_S\rVert_F
\right)
\lVert\Delta x_S\rVert_2.
\end{align}
Applying Lemma~\ref{lem:block_norm_bound} yields
\[
\lVert\widehat y_S-y\rVert_2
\leq
\left[
\sqrt{B}\,c_W
+
\sqrt{B}\,c_A
\left(1+\sqrt{B}\,c_W\right)
\right]
\lVert W_R\rVert_F\lVert x_R\rVert_2.
\]
Squaring and taking the expectation gives
\[
\mathbb{E}\lVert\widehat y_S-y\rVert_2^2
\leq
C\lVert W_R\rVert_F^2
\mathbb{E}\lVert x_R\rVert_2^2.
\]
No channel-independence assumption is required because
\[
\mathbb{E}\lVert x_R\rVert_2^2
=
\sum_{j\notin S}\mathbb{E}[x_j^2]
=
\sum_{j\notin S}m_j.
\]
Finally,
\(
\lVert W_R\rVert_F\leq\lVert W\rVert_F
\),
which proves Equation~\ref{eq:relaxed_upper_bound}.
\end{proof}

The tighter bound in Equation~\ref{eq:tighter_upper_bound} still
depends on \(S\) through \(\lVert W_R\rVert_F\). The relaxation to
\(\lVert W\rVert_F\) is what produces a separable objective over the
activation channels. Then to discuss the optimal selection under a fixed protection budget s, we have the following proposition: 

\begin{proposition}[Optimal selection for the relaxed bound]
\label{prop:top_second_moment}
For a fixed protection budget \(s\), any set containing the \(s\)
largest second moments
\[
m_j=\mathbb{E}[x_j^2]
\]
minimizes the relaxed upper bound in
Equation~\ref{eq:relaxed_upper_bound}. For $1\leq s < K$, the minimizer is unique if 
$m_{(s)}>m_{(s+1)}$. 
\end{proposition}

\begin{proof}
The selection-dependent part of the relaxed bound is
\[
\sum_{j\notin S}m_j
=
\sum_{j=1}^{K}m_j
-
\sum_{j\in S}m_j.
\]
The first term is independent of \(S\). Therefore minimizing the
residual sum is equivalent to maximizing
\(
\sum_{j\in S}m_j
\),
which is achieved by selecting the \(s\) largest second moments. The
selection is unique precisely when there is a strict gap at the
selection boundary.
\end{proof}

However, Proposition~\ref{prop:residual_energy_bound}, \ref{prop:top_second_moment} select channels and bound the quantization error according to the second moments. Therefore, we need to prove that variance ranking agrees with second-moment ranking under suitable conditions. To prove this, we have the following corollary: 

\begin{corollary}[When variance ranking is sufficient]
\label{cor:variance_ranking}
If \(\mu_j=0\) for every channel, then \(m_j=v_j\), and Top-\(s\)
variance minimizes the same relaxed bound.

More generally, let 
$v_{(1)}\geq\cdots\geq v_{(K)}$
be the sorted variances and suppose
\[
\max_j\mu_j^2
<
v_{(s)}-v_{(s+1)}.
\]
Then the Top-\(s\) variance set is also the unique Top-\(s\)
second-moment set.
\end{corollary}

\begin{proof}
If \(\mu=0\), then \(m_j=v_j\) for every \(j\), and the claim follows
from Proposition~\ref{prop:top_second_moment}.

In the general case, the hypothesis forces \(v_{(s)}>v_{(s+1)}\), so
the Top-\(s\) variance set \(S_v\) is well defined. Let \(i\in S_v\)
and \(j\notin S_v\). Then \(v_i\geq v_{(s)}\) and
\(v_j\leq v_{(s+1)}\), so
\[
m_i=v_i+\mu_i^2\;\geq\;v_{(s)}
\;>\;v_{(s+1)}+\max_k\mu_k^2
\;\geq\;v_j+\mu_j^2=m_j.
\]
Every channel in \(S_v\) therefore has a strictly larger second moment
than every channel outside it. Hence \(S_v\) is the unique Top-\(s\)
second-moment set, and \(m_{(s)}>m_{(s+1)}\).
\end{proof}

Finally, we can conclude the \textbf{main theorem.}

\begin{theorem}[Top-\(s\) variance protection bounds the W4A4 error]
\label{thm:variance_protection}
Consider the layer \(y=Wx\) with NVFP4 block width \(B\) and a
protected set \(S\) with \(\lvert S\rvert=s\), served by an exact BF16
side branch as in Equation~\ref{eq:protected_output}. Assume that,
after masking, every activation and weight block scale is an E4M3
normal number that does not saturate. Then the following hold.
\begin{enumerate}
  \item[\textup{(i)}] For every \(S\),
  \(
  \mathbb{E}\lVert\widehat y_S-y\rVert_2^2
  \le C\lVert W\rVert_F^2\sum_{j\notin S}m_j
  \),
  
  with $C = \left[ \sqrt{B}\,c_W + \sqrt{B}\,c_A \left(1+\sqrt{B}\,c_W\right)\right]^2.$ $c_A,c_W$ are the constants in bounding the per-element or per-weight block NVFP4 error, respectively. 
  
  \item[\textup{(ii)}] The relaxed bound is minimized by the Top-\(s\)
  second-moment set. Crucially, if \(\mu=0\), the Top-\(s\) variance set
  minimizes the relaxed bound,  
  so protecting the Top-\(s\) variance channels guarantees
  \[
  \mathbb{E}\lVert\widehat y_S-y\rVert_2^2
  \le
  C\lVert W\rVert_F^2
  \Bigl(\sum_{j=1}^{K}m_j-\sum_{i=1}^{s}m_{(i)}\Bigr),
  \]
  which is the smallest value of the relaxed bound attainable with
  budget \(s\).
\end{enumerate}
\end{theorem}

\begin{proof}
Part~(i) is Proposition~\ref{prop:residual_energy_bound}.
Part~(ii) is
Proposition~\ref{prop:top_second_moment} and 
Corollary~\ref{cor:variance_ranking}, identifying the Top-\(s\) set.  
Substituting that set into
part~(i) gives the displayed inequality, and part~(ii) gives its optimality.
\end{proof}

\paragraph{Interpretation and scope.}
\label{sec:scope}
With the layer-Norm, the condition that $\mu = 0$ can be approximate. Then the theoretical result supports the following interpretation: 

\begin{quote}
High-energy activation channels amplify weight-quantization
perturbations. Moving such channels to a BF16 side branch removes their
direct FP4 product error and can reduce the local K16 scale of the
remaining activation and weight blocks. Under a low-mean or
ranking-preservation condition, activation variance is a computationally
cheap proxy for the relevant second moment.
\end{quote}

\end{document}